\documentclass{article} %
\usepackage[final]{colm2025_conference}
\usepackage{mathtools}
\usepackage{microtype}
\usepackage{hyperref}
\usepackage{url}
\usepackage{wrapfig}
\usepackage{multirow}
\usepackage{booktabs}
\usepackage{graphicx}
\usepackage{subcaption}
\usepackage{amsmath}
\usepackage{lineno}

\usepackage{amsmath,amsfonts,bm}

\def\eqref#1{equation~\ref{#1}}

\def\1{\bm{1}}

\def\vw{{\bm{w}}}
\def\vx{{\bm{x}}}

\DeclareMathAlphabet{\mathsfit}{\encodingdefault}{\sfdefault}{m}{sl}
\SetMathAlphabet{\mathsfit}{bold}{\encodingdefault}{\sfdefault}{bx}{n}

\definecolor{darkblue}{rgb}{0, 0, 0.5}
\hypersetup{colorlinks=true, citecolor=darkblue, linkcolor=darkblue, urlcolor=darkblue}

\usepackage{listings}
\usepackage{xcolor}

\lstdefinelanguage{Prompt}{
    morekeywords={question, options, answer, tag, Example, correct, past, future, irrelevant},
    sensitive=false,
    morecomment=[l]{//},
    morestring=[b]',
}

\lstdefinestyle{promptstyle}{
    language=Prompt,
    basicstyle=\footnotesize\ttfamily,
    keywordstyle=\color{blue},
    stringstyle=\color{teal},
    commentstyle=\color{gray},
    breaklines=true,
    breakatwhitespace=true,
    showstringspaces=false,
    frame=single,
    rulecolor=\color{black},
    columns=flexible
}

\title{TiMoE: Time-Aware Mixture of Language Experts}

\author{Robin Faro$^\star$, Dongyang Fan$^\star$, Tamar Alphaidze, Martin Jaggi \\
EPFL, Switzerland \\
\texttt{firstname.lastname@epfl.ch}}

\begin{document}

\ifcolmsubmission
\linenumbers
\fi

\maketitle

\begin{abstract}
Large language models (LLMs) are typically trained on fixed snapshots of the web, which means that their knowledge becomes stale and their predictions risk \emph{temporal leakage}: relying on information that lies in the future relative to a query.  We tackle this problem by pre-training \emph{from scratch} a set of GPT-style experts on disjoint two-year slices of a 2013–2024 corpus and combining them through \textbf{TiMoE}, a \underline{Ti}me-aware \underline{M}ixture \underline{o}f Language \underline{E}xperts.  At inference time, TiMoE masks all experts whose training window ends after the query timestamp and merges the remaining log-probabilities in a shared space, guaranteeing strict causal validity while retaining the breadth of multi-period knowledge.  We also release \textsc{TSQA}, a 10k-question benchmark whose alternatives are explicitly labelled as \emph{past}, \emph{future} or \emph{irrelevant}, allowing fine-grained measurement of temporal hallucinations.  Experiments on eight standard NLP tasks plus \textsc{TSQA} show that a co-adapted TiMoE variant matches or exceeds the best single-period expert and cuts future-knowledge errors by up to 15\,\%.  Our results demonstrate that modular, time-segmented pre-training paired with causal routing is a simple yet effective path toward LLMs that stay chronologically grounded without sacrificing general performance much. We open source our codes at \href{https://github.com/epfml/TiMoE}{TiMoE (Github)}.
\end{abstract}

\section{Introduction}

Large Language Models (LLMs) have demonstrated remarkable capabilities across a wide range of tasks and domains. However, they still lack a robust understanding of time~\citep{ge-etal-2024-time, mousavi-etal-2024-dyknow, chenghaozhu-etal-2025-llm}. For instance, on the \textsc{FreshQA} benchmark~\citep{vu-etal-2024-freshllms}, GPT-4 achieves only 14\% accuracy on the “fast-changing” subset. This temporal unawareness undermines the reliability of LLMs—particularly in critical fields like medicine, where clinical guidelines, drug approvals, and frontline therapies evolve rapidly. In such contexts, relying on outdated information can mislead patients, resulting in inappropriate treatments or the misuse of antibiotics.

While post-hoc verification with external tools or retrieval can mitigate some issues, it is preferable to embed time-awareness directly in the model’s parameters. Standard pretraining, however, typically merges data from all time periods into a single model indiscriminately. This conflation of temporal contexts makes it difficult for the model to distinguish between what \emph{was} true, what \emph{is} true, and what \emph{will} be true—ultimately compromising its ability to make accurate and time-consistent predictions in dynamic domains.

To address this, we propose a new pretraining strategy that \emph{respects the temporal dimension of language data}. Specifically, we partition the training corpus into non-overlapping time slices and train separate LLMs—referred to as \emph{time-specific experts}—on each slice. This design ensures that each expert learns to model language within a well-defined temporal context, without any contamination from the future. Moreover, another advantage of this approach is its natural support for time-evolving model development. New experts can be added for future time periods simply by training on the corresponding data slices, without requiring any modification to the already trained experts from earlier years. This modularity prevents catastrophic forgetting~\citep{luo2025empiricalstudycatastrophicforgetting}, as past experts are frozen and preserved, allowing the model to retain a faithful representation of historical linguistic patterns. It also facilitates efficient updates over time, enabling the system to continuously adapt to new information and language use without reprocessing the entire dataset or compromising temporal integrity.

The time-specific experts are trained independently to preserve temporal locality and prevent future information leakage. However, at inference time, relying solely on a single expert from one time slice is overly restrictive and may discard relevant historical context. To address this, we propose a \emph{time-aware aggregation mechanism} that conditions on the input timestamp and activates all experts whose training data precedes or aligns with it. This approach introduces a causal mask over the experts, allowing the model to access all available knowledge up to (but not beyond) the query time. 

Together, these components form our time-sensitive pretraining pipeline for language models. The resulting models exhibit significantly improved temporal awareness compared to state-of-the-art LLMs of similar scale, particularly in tasks requiring sensitivity to evolving facts. Beyond performance gains, the modular nature of the pipeline enables systematic analysis of knowledge change over time: by comparing the predictions or representations of different time-specific experts, one can trace how language use, entity relevance, or factual content shifts across periods. This opens up new opportunities for diachronic linguistics, media forensics, and the detection of temporal inconsistencies in both model outputs and training corpora.

Overall, our contributions can be summarized as follows:
\begin{itemize}
    \item We present the first \emph{from-scratch} time-segmented pre-training pipeline for LLMs that enforces strict temporal isolation between periods, eliminating future leakage by design.
    \item We propose \textbf{TiMoE}, a simple yet effective log-probability mixture that (a) respects causal masking and (b) supports both fixed and learned routing strategies.
    \item We curate \textsc{TSQA}, a 10k-question multiple-choice benchmark that tags each distractor as \emph{past}, \emph{future}, or \emph{irrelevant}, enabling fine-grained diagnosis of temporal hallucinations. The benchmark is open sourced at \href{https://huggingface.co/datasets/robinfaro/TSQA}{TSQA (Hugging Face)}.
    \item Empirically, \textsc{TiMoE-CoAdapt} matches or exceeds the best single-period expert on eight standard NLP tasks, while reducing temporal-drift errors on \textsc{TSQA} by up to 15\,\%.
\end{itemize}

\begin{figure}[t!]
    \centering
    \includegraphics[width=\linewidth]{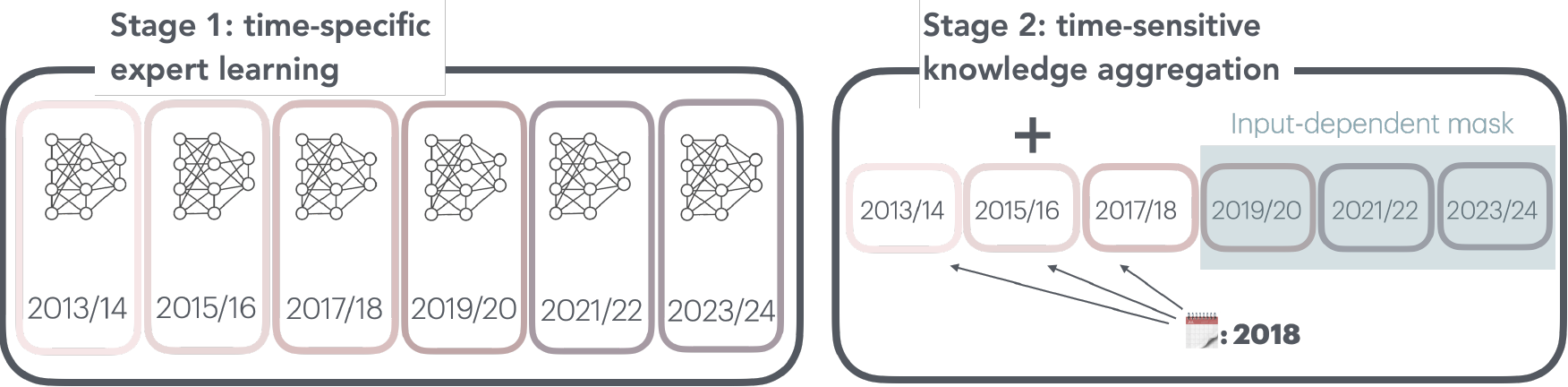}
    \caption{Diagram of our proposed training framework.}
    \label{fig:approach-diagram}
\end{figure}

\section{Related Work}

\paragraph{Time aware pretraining.}
Several works inject \emph{structured time tokens} into the input sequence so that the transformer can learn a dedicated temporal embedding space.  
\citet{Dhingra_2022} prepend a timestamp to every span during T5 pre-training, enabling the model to condition generation on the creation date and to be quickly refreshed with additional dated corpora.  
Similarly, \citet{rosin2022timemasking} introduce \textsc{TempoBERT}, which masks the timestamp itself during masked-language modelling, encouraging stronger temporal representations and yielding sizeable gains on sentence-dating and semantic-shift benchmarks.  Apart from introducing a timestamp token, there is another line of research on designing more time-aware training. \citet{loureiro-etal-2022-timelms} keep a RoBERTa backbone in continual pre-training on monthly Twitter snapshots, mitigating catastrophic forgetting while maintaining freshness on out-of-distribution tweets.  
A complementary strategy is \emph{temporal adaptation}: train a separate adapter for each era.  
\citet{rottger-pierrehumbert-2021-temporal-adaptation} show that BERT variants adapted to year-specific social-media slices outperform a monolithic model on both MLM and downstream hate-speech classification for the target period.  
More recently, \citet{fittschen2025pretraininglanguagemodelsdiachronic} demonstrate that fine-tuning Llama-3-8B on just five 10 M-word historical slices suffices to capture diachronic lexical change with minimal compute. However, adapter-based methods risk temporal leakage, as strong pretrained backbones may retain knowledge across time slices. In contrast, our work performs time-aware pretraining \emph{from scratch}, ensuring strict temporal isolation and eliminating future information leakage.

\paragraph{Benchmarks for time-sensitive knowledge.}
A growing suite of datasets explicitly probes how well language models handle facts that change over time.
\textsc{FreshQA} (\citealt{vu-etal-2024-freshllms}) separates \emph{fast-}, \emph{slow-}, and \emph{never-changing} questions, revealing sharp accuracy drops on the first category.
\textsc{TimeShift} \citep{herel2025timeawarenesslargelanguage} contains 8 000 dated facts (2018-2024) and measures pure recall as the query date moves beyond a model’s training cut-off.
Continuous-update leaderboards such as \textsc{RealTimeQA}~\citep{kasai2023realtimeQA} and \textsc{Daily Oracle}~\citep{dai2024llmsprescientcontinuousevaluation} publish fresh queries weekly or even daily, enabling regression tracking of temporal drift.
More recent work broadens the lens: \textsc{UnSeenTimeQA} \citep{uddin2024unseentimeqatimesensitivequestionansweringllms} stresses counterfactual reasoning over synthetic timelines; \textsc{TimE} \citep{wei2025timemultilevelbenchmarktemporal} and \textsc{TimeBench} \citep{chu2024timebenchcomprehensiveevaluationtemporal} decompose temporal competence into subtasks such as date arithmetic, interval ordering and commonsense timelines; while \textsc{Test of Time} (ToT) \citep{fatemi2025test} cleanly separates semantic ordering from numeric date reasoning. Motivated by these efforts, we introduce a new time-sensitive QA benchmark that categorizes answers as \emph{current}, \emph{past}, \emph{future}, or \emph{irrelevant}, enabling fine-grained analysis of models’ temporal grounding and their propensity to rely on outdated or anticipatory information.

\paragraph{Combination of language models.} \textsc{Model Soups}~\citep{modelsoups} first showed that simple weight–averaging of multiple fine-tuned checkpoints can yield “soups’’ that outperform every constituent model, a result extended to LLMs by \textsc{Fisher Merging}~\citep{fisher-merging} and the recent \textsc{Activation-Informed Merging}~\citep{nobari2025activationinformedmerginglargelanguage}. \textsc{Branch-Train-Merge}~\citep{li2022branchtrainmergeembarrassinglyparalleltraining} is another popular method for merging LLMs in the parameter space, where the authors explored uniform, argmax, and posterior mix strategies. In the realm of parameter-efficient adaptation, \textsc{AdapterFusion}~\citep{pfeiffer-etal-2021-adapterfusion},  \textsc{Compacter}~\citep{compactor2021} and \textsc{Delta-LoRA}~\citep{zi2023deltalorafinetuninghighrankparameters} demonstrate that multiple task-specific adapters can be combined without re-training the backbone.  Dedicated LoRA-composition work—including , \textsc{Merging LoRAs like LEGO}~\citep{zhao2025merging}, and the expert-routed \textsc{MoE-MLoRA}~\citep{yaggel2025moemloramultidomainctrprediction} and \textsc{LoRA Hub}~\citep{huang2024lorahub}—explores fine-grained mixing and routing of LoRA experts to enable modular skill transfer. In this work, we merge knowledge from multiple time-specific experts. While semantic-based routing plays a key role, we also emphasize preventing information leakage from the future.

\section{Our Design}

To achieve a performant and time-aware LM, we propose a two-stage pretraining procedure: 1) time-specific experts learning, where the pretraining corpus is divided by time periods and a separate model is trained for each; 2) time-aware knowledge aggregation, where we learn to combine knowledge from all relevant past periods to represent each document up to its timestamp.

\subsection{Stage 1: Time-specific LM experts Learning}

We consider a temporal range spanning from 2013 to 2024, dividing it into non-overlapping consecutive two-year intervals, resulting in six time-expert LLMs to train. We choose a two-year granularity as a compromise between temporal resolution and data availability, ensuring that each expert is trained on a sufficiently large and coherent corpus while maintaining clear temporal separation.

\subsection{Stage 2: Time-aware aggregation of LM experts}
Our setup raises the question of how to effectively combine the outputs of the time-specialized models at inference time. Since each expert is relatively lightweight, we aim to make them collaborate to produce high-quality predictions by jointly leveraging their temporally grounded knowledge while preserving consistency with the temporal context of the input.

To address this, we merge the outputs of the time-specific language models $(\{E_k\}$, $k$ denotes time periods) by combining their predictions in the log-probability space. This aggregation mechanism maintains probabilistic coherence while integrating information from multiple temporal sources. The \textit{time-awareness} of our approach derives from a temporal masking strategy: for a query related to a specific year $t_q$, we exclude all experts trained on future data (i.e., any expert $E_k$ trained on documents with $k > t_q$). Only experts with training periods $k \leq t_q$ are allowed to participate in the aggregation, ensuring causal validity and preventing information leakage. This setup enables us to enhance predictions by accumulating knowledge from the past, which is especially useful for time-invariant facts or slow-changing trends.

Let $\mathcal{E}(t_q) \coloneqq \{E_k\}_{k \leq t_q}$ be the set of eligible experts for a query timestamp $t_q$. For each expert $E_k \in \mathcal{E}(t_q)$, we denote its log-probability output as $\log P_k(x_{t+1} \mid \vx_{1:t})$, where $x$ denotes the tokens and $t$ denotes relative position of the token in a sequence. For any aggregation weights $\vw_{k}$, the aggregated log-probability can be calculated as:

\begin{equation}
\log P(x_{t+1} \mid \vx_{1:t}) = \log \left( \sum_{E_k \in \mathcal{E}(t_q)} w_k(x) \cdot \exp\left( \log P_k(x_{t+1} \mid \vx_{1:t}) \right) \right)
\end{equation}

where $w_k(x)$ are non-negative weights summing to 1, i.e., $\sum_{E_k \in \mathcal{E}(t_q)} w_k(x) = 1$. We propose the following distinct strategies for defining these weights:

\textbf{TiMoE-Avg.} This approach assigns equal weight to all experts considered up to the date of the given document:

\[
w_k(x) = \frac{1}{|\mathcal{E}(t_q)|} \quad \forall E_k \in \mathcal{E}(t_q)
\]
It requires no additional training and represents a simple ensemble of past experts.

\textbf{TiMoE-LearnedAvg.} Instead of naive averaging, a trainable router network learns to assign weights $w_k(x)$ based on the input $x$. The experts remain frozen during training. The router learns to softly select the most relevant experts given the query content.

\textbf{TiMoE-CoAdapt.} Both the router and the experts corresponding to the training time window are jointly updated. Specifically, for each training document with timestamp $t_d$, we unfreeze only the expert $E_{t_d}$ and keep all other experts frozen. This co-adaptation enables experts to refine their outputs in the context of other collaborating experts, improving their alignment and complementarity without violating temporal constraints.

\textbf{TiMoE-Year.} Additionally, as a baseline, we include a hard-coded routing strategy that always selects the expert $E_k$ whose training time window exactly matches the query year $t_q$. This produces:

$$
\log P(x_{t+1} \mid \vx_{1:t}) = \log P_k(x_{t+1} \mid \vx_{1:t}) \quad \text{with} \quad k = t_q
$$

\section{Experiments}

\subsection{Experimental setup}
\paragraph{Model.} We choose GPT2\footnote{\url{https://github.com/karpathy/nanoGPT}} as our base model architecture. Each of the LM expert employs a GPT2-medium model, with 480M parameters. The context window is 1024. We apply warmup-stable-decay (WSD) scheduler~\citep{hu2024minicpm}, with a max learning rate of $10^{-4}$.

\paragraph{Pretraining datasets.} For pretraining the language models, we use FineWeb-Edu~\citep{fineweb_edu}, a publicly available dataset containing 1.3 trillion tokens from a diverse set of quality-filtered web documents. We utilize an already available 100 billion token sample from this corpus and divide it into six time-based bins according to the documents' crawl timestamps, which range from 2013 to 2024. The token distribution across these time intervals is as follows:

\begin{table}[h!]
\centering
\resizebox{.9\linewidth}{!}{%
\begin{tabular}{lcccccc}
\toprule
\textbf{Year Interval} & \textbf{2013–2014} & \textbf{2015–2016} & \textbf{2017–2018} & \textbf{2019–2020} & \textbf{2021–2022} & \textbf{2023–2024} \\
\midrule
\textbf{Total Tokens}  & 8.7 B & 15.7 B & 24.8 B & 21.3 B & 19.6 B & 9.8 B \\
\bottomrule
\end{tabular}%
}
\label{tab:token_distribution}
\end{table}

In terms of data processing, we concatenate tokens of documents from the same time bin into a single sequence and then split the resulting long sequence into fixed-length sequences of 1024 tokens. An end-of-text token is added between different documents to mark their boundaries. It's important to note that the crawl timestamp often does not reflect the original creation time of a document, meaning each bin may include older content. While this method is not ideal, it effectively prevents future knowledge leakage—an essential requirement for time-aware modeling.

\paragraph{Expert aggregation strategies.} For the aggregation strategies that require a router, we introduce a router that is simply a one-layer MLP, of which the input is of the size of embeddings and the output is the weights to aggregate the time experts. A time-dependent mask is applied to the router.
For stage-2 training, we continue using the \textit{FineWeb-Edu} dataset introduced in the pretraining setup. For \textsc{TiMoE-LearnedAvg}, we train only the router (keeping all experts frozen) using 200 million tokens and a constant learning rate of $10^{-5}$.

For \textsc{TiMoE-CoAdapt}, we update the router and the "present" expert for each input sequence. For example, for an input sequence from the year 2022, we unfreeze the 2021-22 expert and update the weights together with the learnable router. In this case, we use 1 billion tokens and a constant learning rate of $10^{-4}$.

\paragraph{Evaluation metrics.} To assess the performance of our models on general-purpose downstream tasks, we rely on a diverse set of benchmarks: \textsc{Arc-Easy}~\citep{arc-easy}, \textsc{HellaSwag}~\citep{zellers2019hellaswag}, \textsc{Lambada}~\citep{lambada}, \textsc{MMLU}~\citep{mmlu}, \textsc{PIQA}~\citep{piqa}, \textsc{SCIQ}~\citep{sciq}, \textsc{Social-IQA}~\citep{social_iqa}, and \textsc{Winogrande}~\citep{sakaguchi2019winogrande}. These benchmarks assess the models’ capacity for general knowledge understanding and commonsense reasoning.

\subsection{A new curated benchmark for time sensitive knowledge}
\begin{wrapfigure}{rt}{0.4\textwidth}
\vspace{-0.6em}
  \begin{center}
    \includegraphics[width=\linewidth]{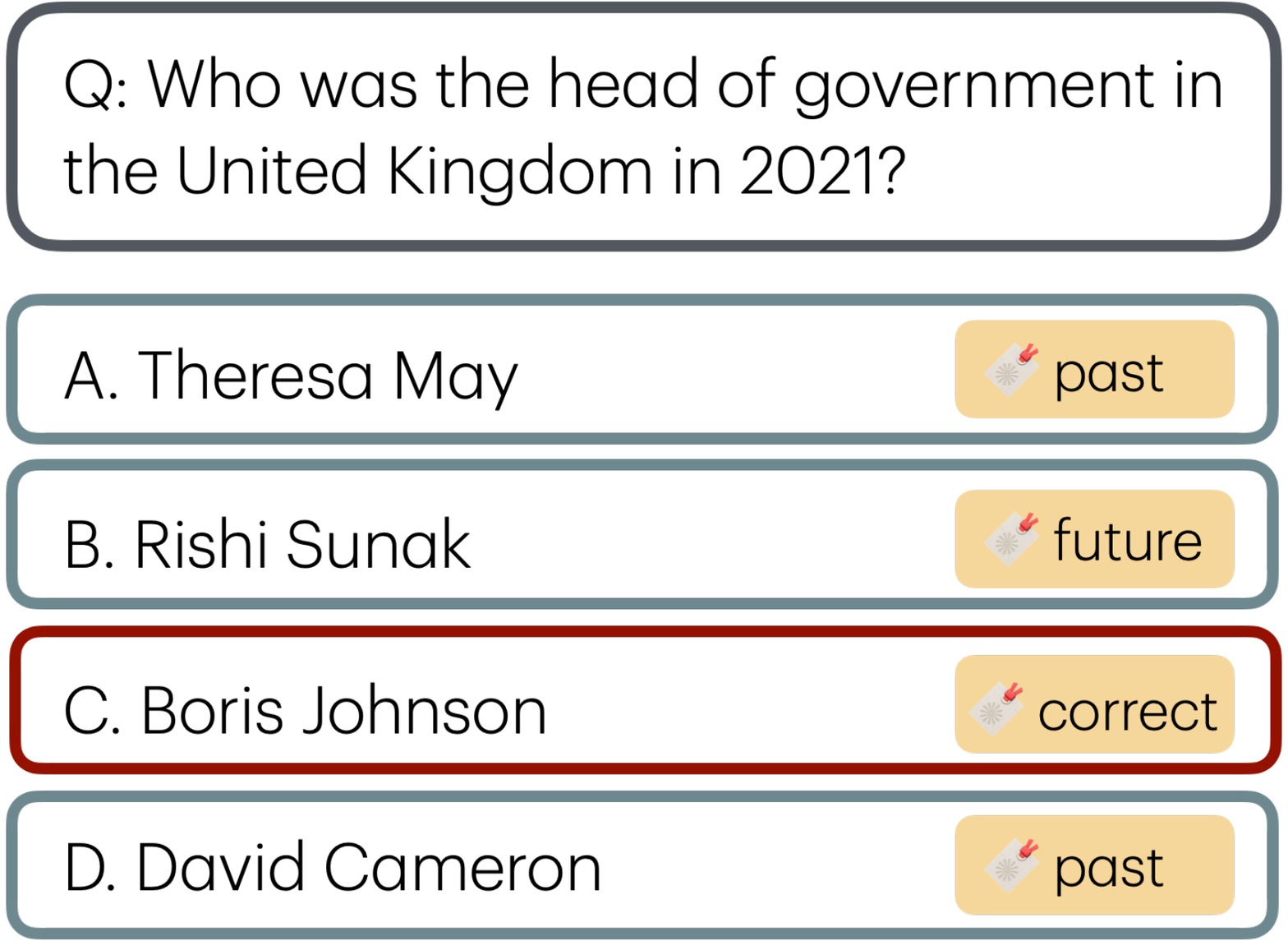}
  \end{center}
  \caption{An example of our TSQA benchmark.}
  \label{fig: tsqa-example}
\end{wrapfigure}

In addition to standard benchmarks, we curate another benchmark (\textsc{TSQA}) to test time-aware factual knowledge. The benchmark spans the years 2013 to 2024, same as the range of our pretraining corpus. In total, it comprises 10{,}063 questions covering events whose correct answer varies over time. Each question is associated with four candidate answers, only one of which is valid for the target year. The remaining options are drawn from other temporal contexts—either from the past, the future, or are irrelevant. An instance of our TSQA is given in Figure~\ref{fig: tsqa-example}.

To construct this benchmark, we start from three existing datasets of temporally grounded factual knowledge: \textsc{DyKnow}~\citep{mousavi-etal-2024-dyknow}, \textsc{Time-Sensitive QA}~\citep{chen2021datasetansweringtimesensitivequestions}, and \textsc{Temporal Alignment QA}~\citep{zhao2024setclocktemporalalignment}. For each event, we extract a timeline of answer variants across different years. We then prompt a DeepSeekV3~\citep{deepseekai2025deepseekv3technicalreport} model to generate a four-choice question for each year: the correct answer is selected from the event's timeline for the target year, while the remaining options are drawn from the answers corresponding to other years and labeled as \textit{past} or \textit{future}. When the timeline does not provide enough temporally grounded alternatives, we supplement them with a plausible but \textit{irrelevant} answer generated by the LLM.

Since our models are pre-trained but not instruction-tuned for high-level interaction, we evaluate them using a standard multiple-choice format consistent with standard libraries for benchmarking, e.g. lm-eval-harness~\citep{eval-harness}. Specifically, for each question and its set of answer choices, we compute the log-likelihood of the final token in each candidate answer when conditioned on the query, and select the option with the highest score.

This method evaluates the model's ability to assign higher likelihood to the correct answer purely based on its pretraining, without relying on task-specific tuning or prompting.

\subsection{Comparable performance on general-purpose benchmarks}

We first test our time-aware models on general-purpose benchmarks. As the benchmarks are not time-sensitive, we do not mask out any experts during inference time, i.e. the timestamp is always set as the most recent year (2024). This effectively simulates an up-to-date model. We also evaluate each of the six underlying GPT-2 experts independently and report both the average and the maximum accuracy across them (\textsc{Year (mean)} and \textsc{Year Expert (max)} respectively). Additionally, we include a baseline model, \textsc{GPT2-Full}, which is trained on the aggregated datasets without any temporal segmentation. This serves as a strong non-modular baseline, as the model is trained the most tokens. The results are reported in Table~\ref{tab:benchmarks_results}, where \textsc{TiMoE-CoAdapt} consistently outperforms both the average and the best single-expert baselines across most tasks, demonstrating the effectiveness of our aggregation strategy. Notably, we observe that \textsc{TiMoE-CoAdapt} underperforms the unified \textsc{GPT2-Full} baseline by an average of 5.6\% across tasks, highlighting the "curse of time-awareness".

\begin{table}[ht]
\centering
\caption{
Accuracy on general-purpose benchmarks. GPT2-Full is a non-modular baseline trained on the entire 100B-token corpus and serves as an upper-bound reference. Underlined values indicate the best performance among modular configurations, excluding GPT2-Full, and are used to assess the effectiveness of different expert aggregation strategies.
}
\resizebox{.9\linewidth}{!}{%
\begin{tabular}{lcccccc}
\toprule
\multirow{2}{*}{\textbf{Task}} & \multirow{2}{*}{\textbf{GPT2-Full}} & \multicolumn{5}{c}{\textsc{TiMoE}} \\
& & \textbf{Year (mean)} & \textbf{Year (max)}  & \textbf{Avg} & \textbf{LearnedAvg}& \textbf{CoAdapt}  \\
\midrule
\textsc{arc\_easy}       & \textbf{0.662} & 0.578 & 0.601      & 0.601 & 0.602 & \underline{0.606} \\
\textsc{hellaswag}       & \textbf{0.368} & 0.318 & 0.325      & 0.319 & 0.323 & \underline{0.326} \\
\textsc{lambada}         & \textbf{0.386} & 0.297 & 0.329     & 0.331 & 0.337& \underline{0.344}   \\
\textsc{mmlu}            & \textbf{0.299} & 0.277 & \underline{0.286}  & 0.281 & 0.282 & 0.282 \\
\textsc{piqa}            & \textbf{0.697} & 0.659 & 0.671      & 0.667 & 0.670  & \underline{0.671}\\
\textsc{sciq}            & \textbf{0.860} & 0.808 & 0.836      & 0.838 & \underline{0.843} & \underline{0.843} \\
\textsc{social\_iqa}     & \textbf{0.394} & 0.378 & \underline{0.391}  & 0.389 & 0.388 & 0.387 \\
\textsc{winogrande}      & \textbf{0.532} & 0.510 & 0.519      & \underline{0.520} & 0.519 & 0.513 \\
Avg                      & \textbf{0.525} & 0.478 & 0.495      & 0.493 & \underline{0.496} & \underline{0.496}\\
\bottomrule
\end{tabular}%
}

\label{tab:benchmarks_results}
\end{table}

\subsection{Improved time awareness through time-sensitive modeling}
We evaluate our models on our curated \textsc{TSQA} benchmark. For our family of \textsc{TiMoE} models, the timestamp is explicitly provided as an additional input parameter to the model, enabling time-awareness when masking out future LM experts. In contrast, for non-time-aware baselines (i.e., \textsc{Falcon 3}~\citep{almazrouei2023falconseriesopenlanguage}, \textsc{Qwen 2.5}~\citep{qwen2025qwen25technicalreport}, \textsc{LLaMA 3.2}~\citep{grattafiori2024llama3herdmodels},
\textsc{GPT2-Full}), the temporal information is explicitly stated within the query prompt. 

\begin{figure}[h]
    \centering
    \includegraphics[width=.9\linewidth]{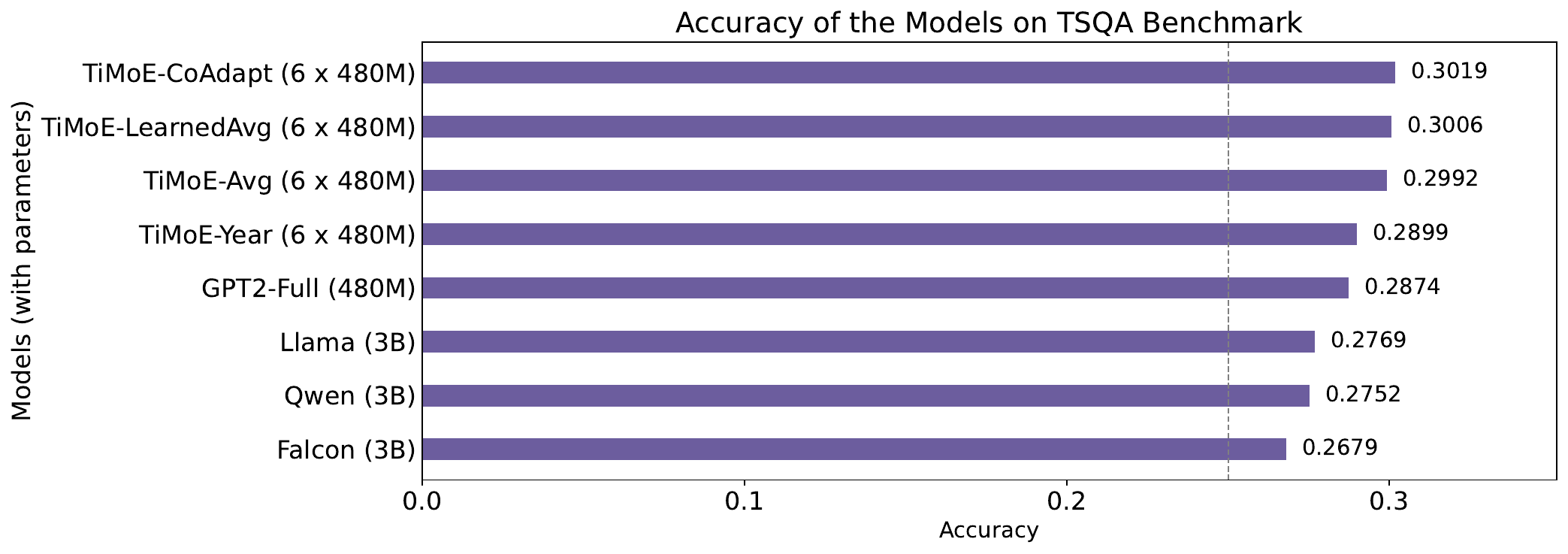}
    \caption{Accuracies of the models on our curated TSQA benchmark}
    \label{fig:accuracy-comparison}
    \vspace{-1em}
\end{figure}

Our primary findings are summarized in Figure~\ref{fig:accuracy-comparison}, which highlights the performance of our proposed TiMoE variants. In particular, the \textsc{TiMoE-LearnedAvg} and \textsc{TiMoE-CoAdapt} models demonstrate a consistent and notable performance advantage of approximately 3\% over frontier language models of comparable size. This is especially impressive considering that these baseline models are built on substantially larger backbones—specifically, 3-billion-parameter architectures—while our TiMoE variants are constructed using significantly smaller 480-million-parameter GPT-2 experts.

\begin{figure}
    \centering
    \includegraphics[width=.9\linewidth]{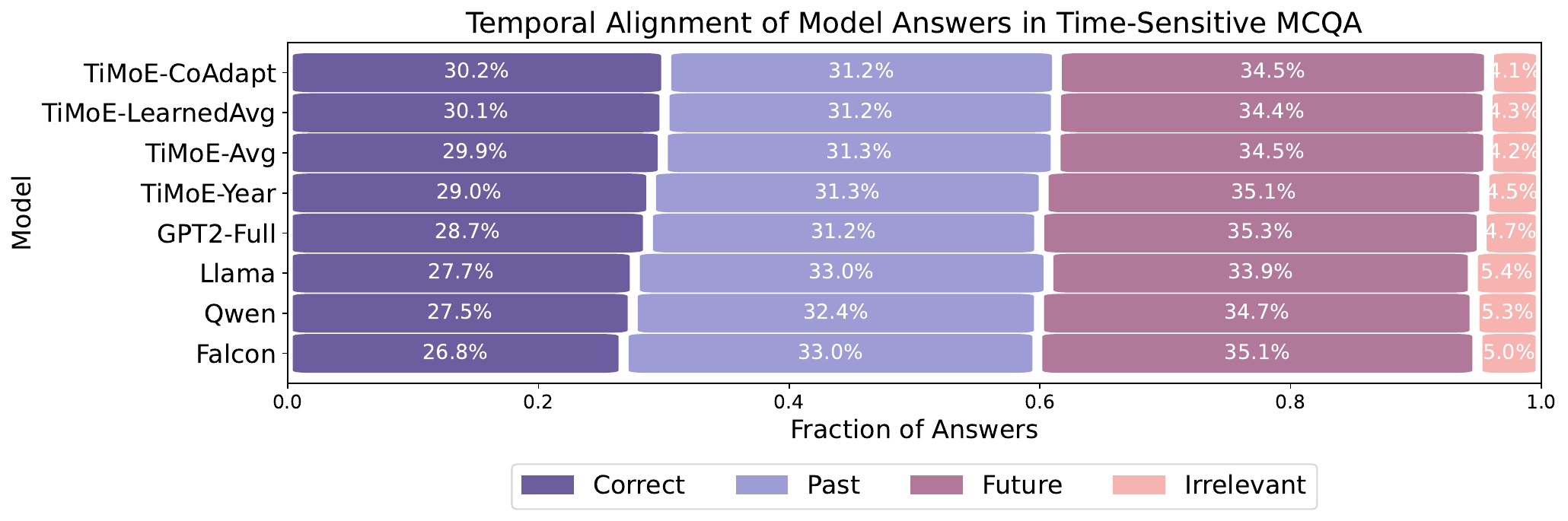}
    \caption{Distribution of answer types (Correct, Temporally Past, Temporally Future, Irrelevant) across evaluated models on our curated TSQA benchmark}
    \label{fig:answer-distribution}
    \vspace{-1em}
\end{figure}

A more detailed examination reveals that our TiMoE models are especially effective at minimizing the inclusion of \emph{past} and \emph{irrelevant} responses, which often dilute the quality of temporal reasoning in language models. This improvement is evident in the answer distribution visualized in Figure~\ref{fig:answer-distribution}, where TiMoE models show a clearer focus on contextually and temporally appropriate outputs.

\begin{figure}[t!]
    \centering
    \includegraphics[width=.9\linewidth]{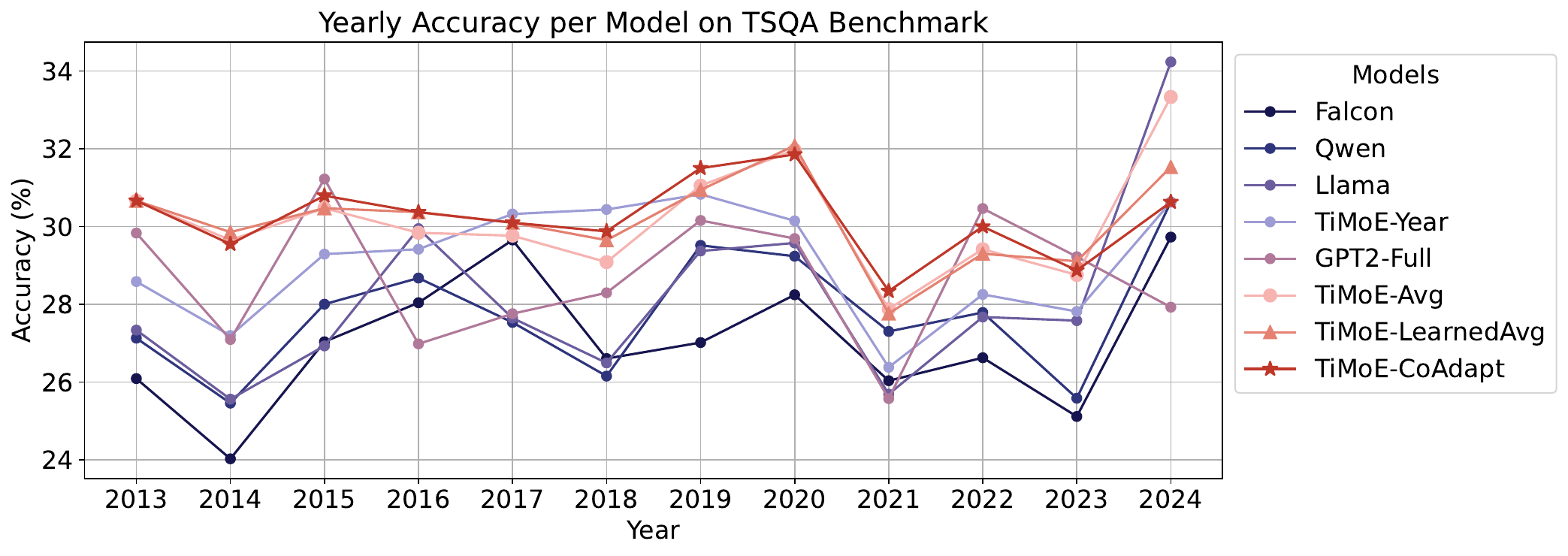}
    \caption{Yearly accuracies of the models on our curated TSQA benchmark 
    }
    \label{fig:yearly-accs}
\end{figure}

Furthermore, the benefits of time-awareness introduced by the TiMoE framework persist across all examined time periods. As illustrated in Figure~\ref{fig:yearly-accs}, our models consistently outperform the baselines year by year, underscoring their robustness and generalizability in temporally dynamic settings.

\subsection{Cross-temporal semantic change discovery}

\begin{figure}[t!]
    \centering
    \includegraphics[width=.7\linewidth]{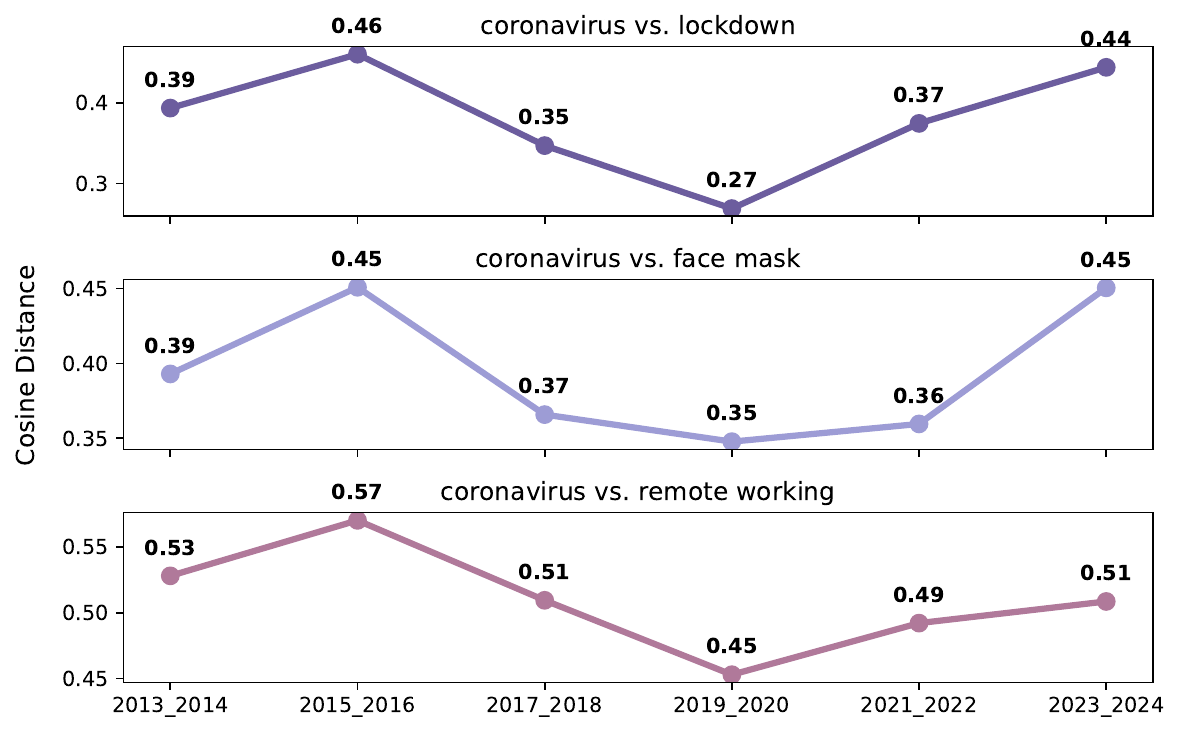}
    \caption{Semantic Distance from word 'Coronavirus' in years 2013-2014 vs. 2023-2024}
    \label{fig:semantic-distance-change}
\end{figure}

Our time-aware models offer a convenient framework for analyzing lexical semantic change over time, as they allow for the extraction of time-specific embeddings from models trained independently on different temporal slices. For our analysis, we extract the embedding of the final token from the last hidden layer, treating it as the latent representation of the input. Following the approach of \citet{periti2024LSC}, we compute the cosine similarity between embeddings of word pairs as a measure of semantic distance, with the changes over time visualized in Figure~\ref{fig:semantic-distance-change}.

We observe that the semantic distances between the term ``coronavirus" and related terms such as ``lockdown", ``face mask", and ``remote working" are smallest in the 2019–2020 model, aligning with the onset of the COVID-19 pandemic during that period, when these concepts were frequently discussed together. In contrast, by 2022–2023, as strict public health measures like lockdowns and quarantines were relaxed, the lexical distances between these terms increased, reflecting a semantic drift in their contextual associations.

\section{Conclusion}
TiMoE demonstrates that partitioning pre-training data into strict time slices and blending the resulting GPT-2 experts through a causal, timestamp-aware router yields language models that stay chronologically grounded without a heavy accuracy penalty. By masking out any expert trained on data newer than the query year, TiMoE eliminates future-knowledge leakage while letting earlier specialists cooperate, cutting temporally inconsistent answers on the new 10 k-question TSQA benchmark by roughly 15\%and delivering steadier accuracy across years. On eight standard NLP benchmarks, our models slightly underperform a standard trained GPT2 model, highlighting a manageable “cost of time-awareness” rather than a fundamental barrier. Overall, TiMoE offers a simple, modular recipe for building LLMs that respect temporal causality and can serve as a foundation for scalable, ever-evolving models and for diachronic analyses of how knowledge shifts over time.

\section{Limitations and Future Work}
The timestamps used to partition the pretraining corpus reflect the date when each document was collected, rather than its original creation date. As a result, each time slice may contain content originating from various earlier periods. Given that we train on a limited number of tokens per slice, this can lead to present-day knowledge being underrepresented in the overall training distribution, making time-sensitive evaluation more challenging.

Looking ahead, it would be valuable to explore finer-grained temporal partitioning, scale the approach to larger base models and partition the pretraining corpus by the creation dates. These directions hold promise for improving the model’s sensitivity to temporal context and its ability to capture evolving language dynamics more precisely.

\paragraph*{Acknowledgement.} We acknowledge funding from SNSF Grant number 10005248.

\newpage

\bibliography{colm2025_conference}
\bibliographystyle{colm2025_conference}

\newpage
\appendix
\section{Appendix}

\subsection*{Prompt Template for Time-Sensitive QA Generation}

\begin{lstlisting}[style=promptstyle]
You are given a question whose answer may vary over time.

Original question:
<question>

Year of interest: <year_of_interest>
Possible answers across years:
<year_1>: <answers_1>
<year_2>: <answers_2>
...

Task instructions:
1. Choose 1 correct answer for the given year, based on the available answers from the dataset.
2. Select 3 different wrong answers by considering the other possible answers from the dataset across different years.
3. If there is not enough context to choose 3 wrong answers based on the timeline, generate the necessary ones (different from the already chosen) and ensure that the wrong answers are still coherent with the available data.
4. If an answer is not related to the timeline for the given question, it should be tagged as 'irrelevant'. Use this tag when there is not enough data from the dataset's timeline to identify a valid answer.
5. Tag each answer with one of the following labels: 'correct', 'past', 'future', or 'irrelevant'. The 'irrelevant' tag should be used if an answer does not match the timeline for the given question or when insufficient data from the timeline exists to make the answer relevant.
6. If the correct answer also appears in the timeline as 'future' or 'past', do **not** include it as a wrong answer. Ensure that it is tagged correctly as 'correct', 'past', or 'future' based on its timeline.
7. Return the question along with 4 options (1 correct and 3 wrong) in the following dictionary format:
   {
       'question': <question>,
       'options': [
           {'answer': <answer 1>, 'tag': <tag 1>},
           {'answer': <answer 2>, 'tag': <tag 2>},
           {'answer': <answer 3>, 'tag': <tag 3>},
           {'answer': <answer 4>, 'tag': <tag 4>}
       ]
   }
Please give only the dictionary and ensure the format is followed strictly.
Do not provide any additional text, symbols, or explanations.
Follow the example closely:

Example:
question: 'What was the role of Karla Estrada in her most recent television series in 2013?'
options: [
    {'answer': 'Apple Puno', 'tag': 'correct'},
    {'answer': 'Carlita Delyon', 'tag': 'future'},
    {'answer': 'Tita Marichris Matahimik', 'tag': 'past'},
    {'answer': 'Quarter 1-4 Judge Tawag ng Tanghalan', 'tag': 'future'}
]
\end{lstlisting}

\end{document}